\documentclass[lettersize,journal]{IEEEtran}
\usepackage{amsmath,amsfonts}
\usepackage{algorithmic}
\usepackage{algorithm}
\usepackage{array}
\usepackage[caption=false,font=normalsize,labelfont=sf,textfont=sf]{subfig}
\usepackage{textcomp}
\usepackage{stfloats}
\usepackage{url}
\usepackage{graphicx}
\usepackage{subcaption}
\usepackage{verbatim}
\usepackage{graphicx}
\usepackage{cite}
\usepackage{svg}
\usepackage{caption}
\usepackage{colortbl}
\usepackage{multirow}
\usepackage{booktabs} 
\usepackage{marginnote}
\usepackage{hhline}
\usepackage{boldline}
\usepackage{xcolor}
\usepackage{tabularx}
\usepackage{multirow}
\usepackage{bigstrut}
\usepackage{subcaption}
\usepackage{amssymb} 

\begin{document}

\title{RS-DFM: A Remote Sensing Distributed Foundation Model for Diverse Downstream Tasks}

\author{Zhechao~Wang, 
	Peirui~Cheng, 
	Pengju~Tian,
	Yuchao~Wang,
	Mingxin~Chen,
	Shujing~Duan,
	Zhirui~Wang, \IEEEmembership{Member,~IEEE},
	Xinming Li, 
	Xian~Sun, \IEEEmembership{Senior Member,~IEEE}
	
	\thanks{This work was supported by the National Natural Science Foundation of China under Grant 62331027 and Grant 62076241.
		\textit{(Corresponding author: Zhirui Wang.)}}
	\thanks{Zhechao Wang, Pengju~Tian, Yuchao~Wang, Mingxin~Chen, and Xian Sun are with the Aerospace Information Research Institute, Chinese Academy of Sciences, Beijing 100190, China, also with the Key Laboratory of Network Information System Technology (NIST), Aerospace Information Research Institute, Chinese Academy of Sciences, Beijing 100190, China, also with the University of Chinese Academy of Sciences, Beijing 100190, China, and also with the School of Electronic, Electrical and Communication Engineering, University of Chinese Academy of Sciences, Beijing 100190, China (e-mail: wangzhechao21@mails.ucas.ac.cn; 
		tianpengju22@mails.ucas.ac.cn; 
		wangyuhcao22@mails.ucas.ac.cn; 
		chenmingxin22@mails.ucas.ac.cn; 
		duanshujing21@mails.ucas.ac.cn;   sunxian@aircas.ac.cn).}
	\thanks{Peirui Cheng, Zhirui Wang and Xinming Li are with the Aerospace Information Research Institute, Chinese Academy of Sciences, Beijing 100094, China, and also with the Key Laboratory of Network Information System Technology (NIST), Aerospace Information Research Institute, Chinese Academy of Sciences, Beijing 100190, China (e-mail: chengpr@aircas.ac.cn;
		zhirui1990@126.com;  13911729321@139.com).}
}

\markboth{Journal of \LaTeX\ Class Files,~Vol.~14, No.~8, August~2021}%
{Shell \MakeLowercase{\textit{et al.}}: A Sample Article Using IEEEtran.cls for IEEE Journals}


\maketitle

\begin{abstract}
Remote sensing lightweight foundation models have achieved notable success in online perception within remote sensing.
However, their capabilities are restricted to performing online inference solely based on their own observations and models, thus lacking a comprehensive understanding of large-scale remote sensing scenarios.
To overcome this limitation, we propose a Remote Sensing Distributed Foundation Model (RS-DFM) based on generalized information mapping and interaction. 
This model can realize online collaborative perception across multiple platforms and various downstream tasks by mapping observations into a unified space and implementing a task-agnostic information interaction strategy. 
Specifically, we leverage the ground-based geometric prior of remote sensing oblique observations to transform the feature mapping from absolute depth estimation to relative depth estimation, thereby enhancing the model's ability to extract generalized features across diverse heights and perspectives.
Additionally, we present a dual-branch information compression module to decouple high-frequency and low-frequency feature information, achieving feature-level compression while preserving essential task-agnostic details. 
In support of our research, we create a multi-task simulation dataset named AirCo-MultiTasks for multi-UAV collaborative observation.
We also conduct extensive experiments, including 3D object detection, instance segmentation, and trajectory prediction. The numerous results demonstrate that our RS-DFM achieves state-of-the-art performance across various downstream tasks.

\end{abstract}

\begin{IEEEkeywords}
	distributed foundation model, multi-platform collaboration, generalized feature mapping, information decoupling.
\end{IEEEkeywords}

\section{Introduction}
Remote sensing foundation models have been widely applied in remote sensing intelligent perception\cite{yan2023air,sun2023single}, relying on powerful generalization performance. Moreover, the constant iteration of remote observation platform computational resources and the rapid development of lightweight techniques, such as knowledge distillation and quantization, have provided the necessary conditions for online inference of remote sensing foundation models. For example, Ringmo-lite is able to achieve online inference for object detection, semantic segmentation and other tasks on a single-edge terminal.

However, remote sensing scenarios typically cover extensive areas and present complex environment structures. The observational data from a single platform are limited by perspectives and resolutions, making it difficult to encompass all information on the observed objects and even facing issues such as environment occlusion and imaging disturbance. Considering the significant correlation and complementarity among the multi-perspective, multi-resolution observation data from distinct observation platforms, which can provide more comprehensive object information, multi-platform collaborative perception is an effective solution to address the limitations of single-platform observations. However, current remote sensing foundation models only perform online perception based on their own observations, failing to fully utilize the advantages of multi-platform collaborative observation and computational resources, making it difficult to achieve online perception based on multi-platform collaborative observation.

Therefore, to fully leverage multi-platform observation data for multi-task online perception towards extensive remote sensing scenarios, it is essential to design a foundation model that supports online collaborative perception across multiple platforms. The model can not only utilize each observation platform's computational resources to achieve online independent perception and analysis but also leverage the observation data and computational resources from different platforms to fully explore the association relationships among multi-platform observation data through multi-platform information extraction and interactive fusion, thereby achieving multi-platform collaborative perception and analysis.

\begin{figure}[t]  
    \centering  
    \includegraphics[width=0.5\textwidth]{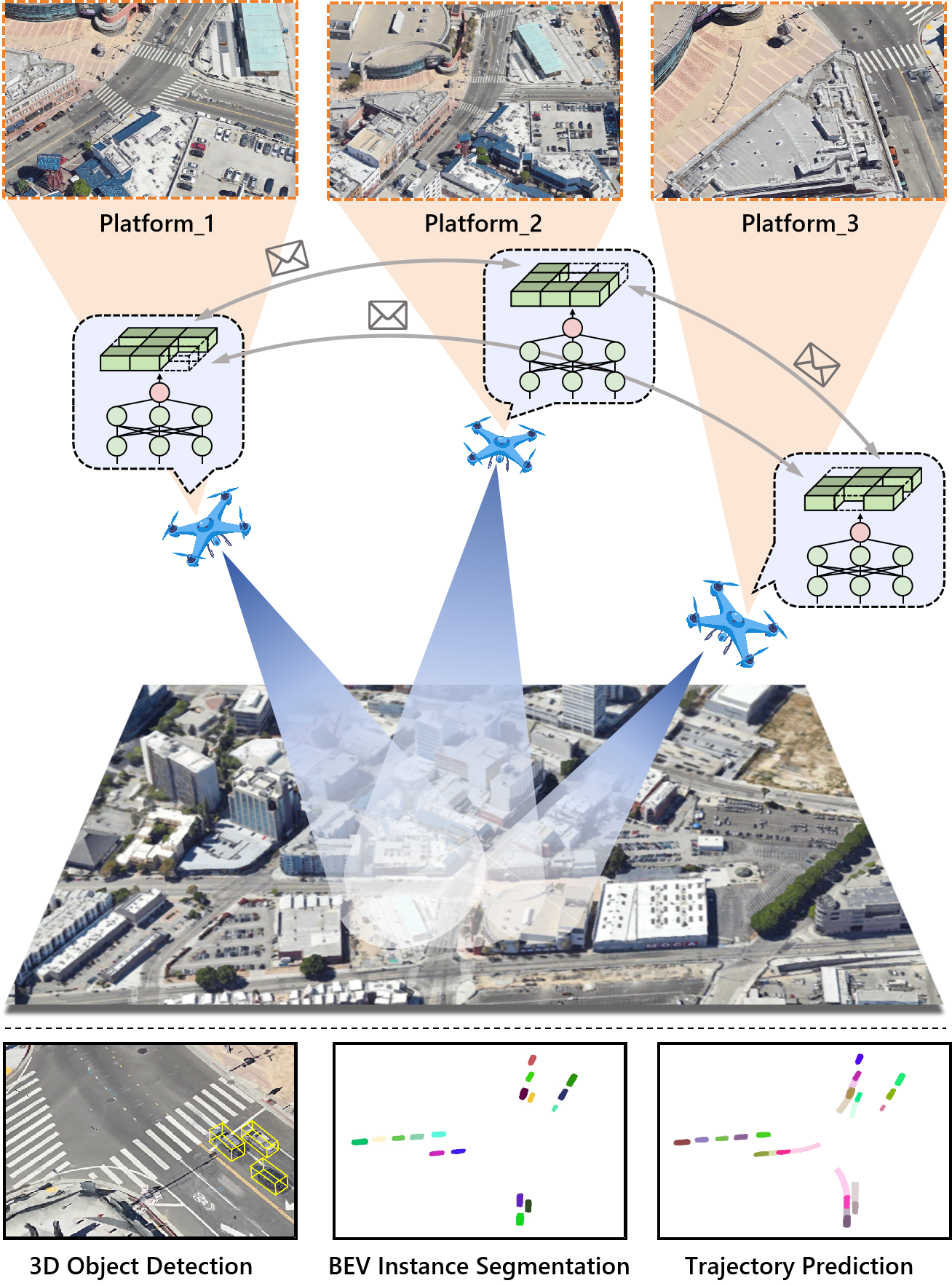}  
    \caption{Illustration of collaborative perception among multiple remote sensing platforms. In this scene, several platforms observe the same area from different angles and heights. Through inter-platform collaboration, they achieve a more comprehensive understanding of the scene, enhancing the performance of various downstream tasks.}  
    \label{intro}  
\end{figure}  
However, constructing an online collaborative perception foundation model across multiple remote sensing platforms presents two significant challenges. The first challenge concerns the consistent recognition issue across distinct platforms. The significant differences in the altitudes and perspectives across various observation platforms lead to inconsistent recognition of information sensitivity to their observations. Therefore, there are severe gaps among multi-platform information, making it challenging to achieve information fusion across platforms and precise collaborative analysis towards objects. Additionally, the information compression loss caused by communication limitations between platforms is the second challenge. Remote sensing observation platforms need real-time observation information interaction to perform online collaborative inference. Remote sensing images have high resolution, and remote sensing platforms are typically distributed over extensive and complex environments, where the wireless communication distance and bandwidth have limitations. Therefore, it is inevitable to compress the information during the real-time information interaction between different observation platforms. However, during information compression, the loss of some critical information results in incomplete object information and even interferes with the observation data itself, thereby reducing the multi-platform collaborative reasoning capability.

To address these challenges, this paper propose a multi-platform collaborative inference framework based on the concept of “physical separation, logical integration”. This framework is enhanced in two key areas: collaborative information extraction and selective information compression interaction. By mapping observational information from different altitudes and perspectives into a unified space, and by decoupling high-frequency and low-frequency information in remote sensing for selective information compression, we achieve online multi-task collaborative perception across multiple remote sensing observation platforms. Specifically, we propose a Remote Sensing Distributed Foundation Model based on information generalized interaction and fusion, which enables individual independent perception on a single terminal and group collaborative perception across multiple platforms. Moreover, we propose a Generalized BEV Generalization (GBG) module, which utilizes the geometric prior of remote sensing observation perspectives to calculate the upper depth limit of pixels. This approach transforms depth estimation from an absolute classification task into a relative classification task based on the upper depth limit, thereby enabling the generalized mapping of different altitudes and perspectives. Finally, we propose a High-Low Frequency Decoupled Collaboration (HLFDC) module. This module leverages distinct attention mechanisms to extract high-frequency and low-frequency features. After employing compression methods for these features, the module transmits the most critical information with minimal bandwidth. To validate the effectiveness of our algorithm, we developed a simulated dataset for multi-UAV collaborative observation, named AirCo-MultiTasks. The dataset contains 5 UAVs at different flight heights and sets rich scenes and objects, supporting multi-tasks such as detection, tracking and prediction. Through experimental validation on this dataset, our algorithm implemented SOTA on various downstream tasks.
\begin{figure*}[t] 
    \centering  
    \includegraphics[width=\textwidth]{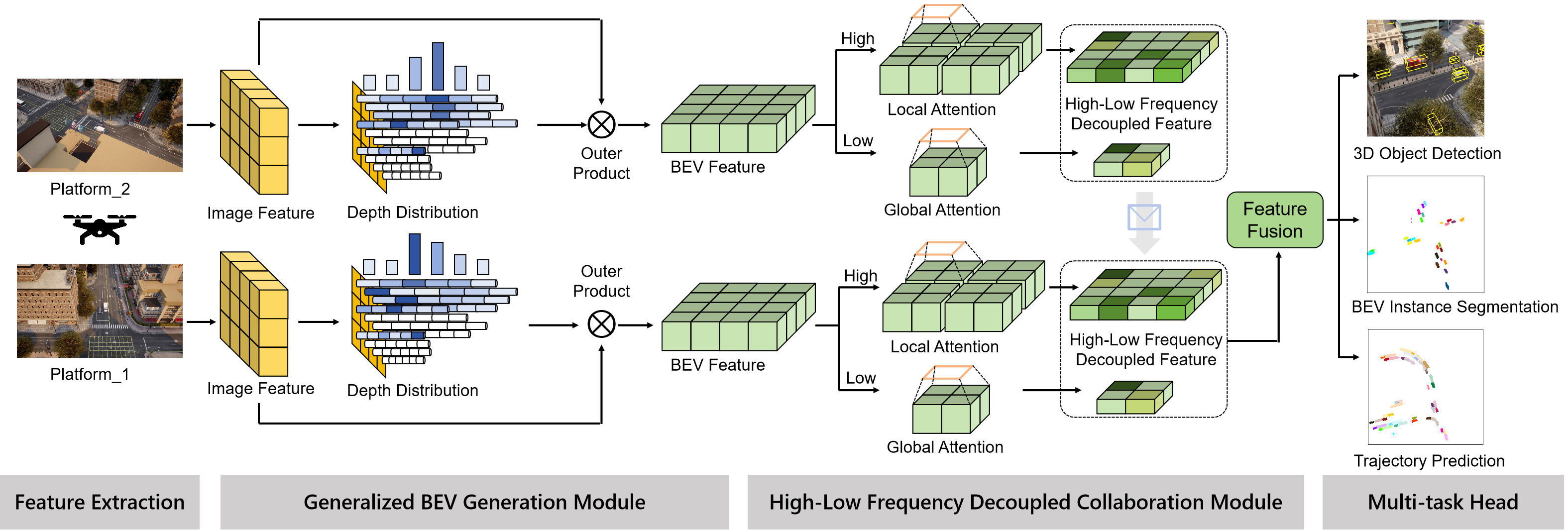} 
    \caption{The overall architecture of our proposed RS-DFM framework. For clarity, we just present the collaboration between two platforms.}  
    \label{overview}  
\end{figure*}  

The main contributions of this paper can be summarized as follows:

We propose a remote sensing distributed foundation model based on information generalized interaction and fusion, which enables individual independent perception on a single terminal and group collaborative perception across multiple platforms.

We present a generalized BEV generation module, which utilizes remote sensing observation geometric prior to achieve the generalized mapping of features from different altitudes and perspectives.

We design a high-low frequency decoupled collaboration module, which reduces transmission bandwidth and achieves efficient selective information transmission.

A large-scale multi-task multi-UAV collaborative perception dataset, AirCo-MultiTasks, is introduced to validate the effectiveness of the proposed framework.

\section{Related Work}
\subsection{Remote Sensing Foundation Model}
The value of using foundation models has become standard practice in computer vision. Many models provide strong baselines and improve performance across various tasks. ViT\cite{dosovitskiy2020image} breaks through the limitations of traditional CNNs by utilizing the Transformer\cite{vaswani2017attention} architecture. MAE\cite{he2022masked} utilizes transformer-based masked image reconstruction approaches to enhance the model's robustness and representation capability and SimMIM\cite{xie2022simmim} further achieves efficient image modeling through a simplified design.

Foundation models have also found widespread applications in intelligent remote sensing perception, with numerous excellent models emerging continuously. RingMo\cite{sun2022ringmo} is the first generative self-supervised foundation model framework in the field of RS. This framework exploits masked image modeling to obtain general feature representation and increase the accuracy of various RS perception tasks. Billionmodel\cite{cha2023billion} introduces the first billion-scale foundation model in the field of remote sensing, marking the largest model size explored in this domain. RingMo-SAM\cite{yan2023ringmo} can process both optical and SAR remote sensing data, capable of zero-shot semantic segmentation and object categories identification, with strong generalization ability.

However, foundation models are not flexible and efficient due to the high demand for computing and storage resources, making it difficult to adapt to edge servers or terminals, and cannot support on-orbit RS image perception in practical applications. RingMo-Lite\cite{wang2024ringmo} aims to address the high computational resource requirements in remote sensing image perception, enabling it to run on edge devices while maintaining high accuracy and versatility across multiple tasks.

\subsection{Collaborative perception}
Collaborative perception is a pivotal approach for intelligent data perception. It is efficiently applied in various scenarios by collaboratively processing information from multiple observation platforms. In autonomous driving, Who2com\cite{liu2020who2com} implements a handshake mechanism to choose collaborative partners, while When2com\cite{liu2020when2com} decides the appropriate time to start collaboration. Where2com\cite{hu2022where2comm} utilizes the detection head to guide regions for sparse interactions. CoCa3D\cite{hu2023collaboration} achieves collaborative depth estimation through considering single-view depth probability and multi-view consistency. HEAL\cite{lu2024extensible} proposes an extensible heterogeneous collaborative perception framework, which ensures extensibility by establishing a unified feature space and aligning new agents to it. CodeFilling\cite{hu2024communication} introduces codebook-based message representation and information-filling-driven message selection for perception-communication trade-off.

Collaborative perception also plays an important role in remote sensing scenarios. MDCNet\cite{duan2024mdcnet} proposes a multi-platform distributed collaborative inference network for object detection. Through the selective communication strategy for multi-platform feature interaction, the method improves the accuracy of object detection and optimizes the inference efficiency by breaking through the limitations of single-platform observation range and precision. DCNNet\cite{zhang2023dcnnet} propose a terminal-cloud collaborative, progressive inference mechanism for image classification, with a distributed self-distillation paradigm designed to integrate and refine in-depth features, performing efficient knowledge transfer between the terminals and the cloud network. 

\begin{figure*}[t] 
    \centering  
    \includegraphics[width=0.65\textwidth]{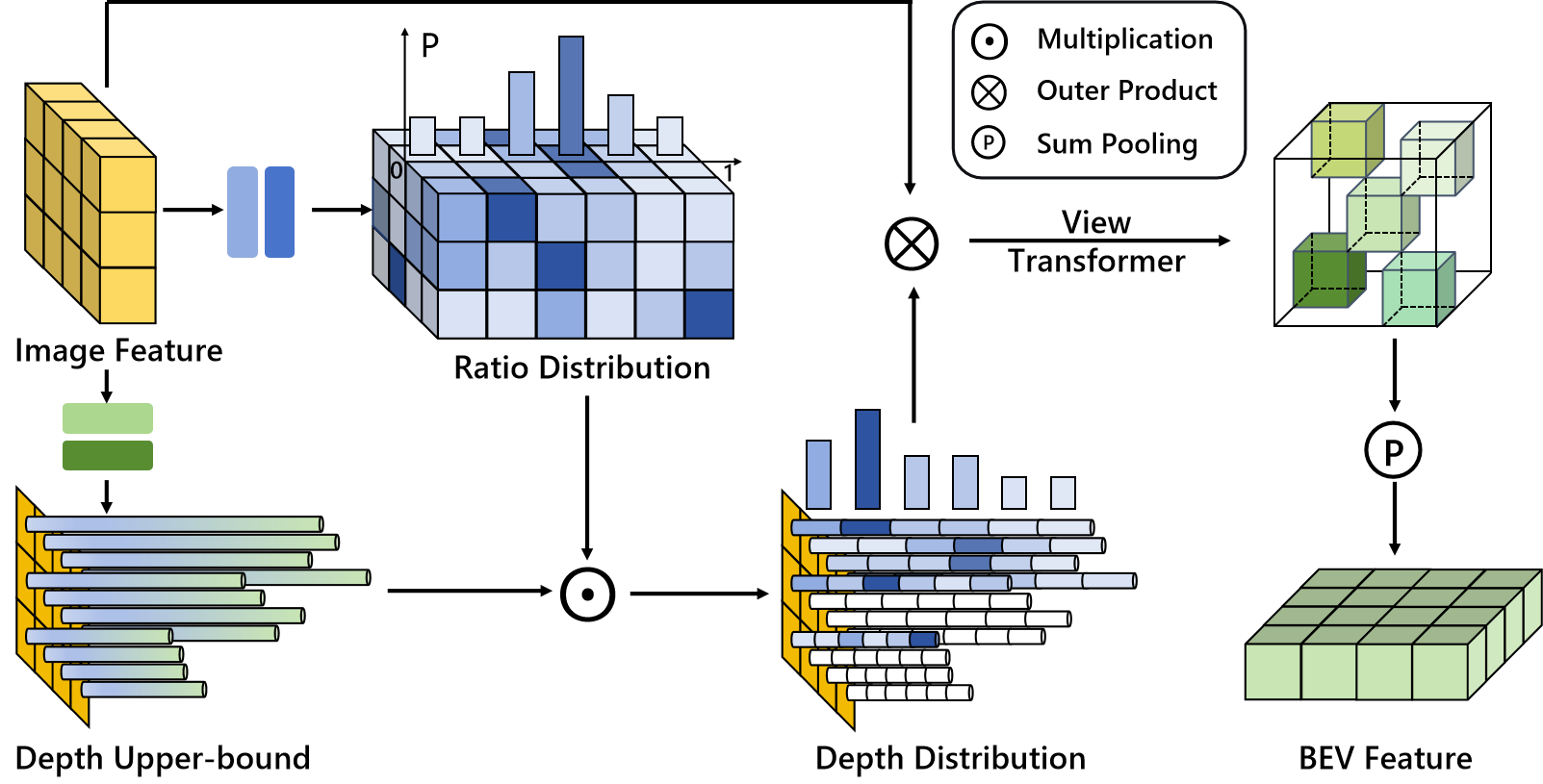} 
    \caption{Operation flow of the GBG module. This module introduces a generalized feature mapping through relative depth estimation, which enhances the accuracy of BEV representation.}  
    \label{gbg}  
\end{figure*}  

\section{Methods}
\begin{figure*}[t] 
    \centering  
    \includegraphics[width=\textwidth]{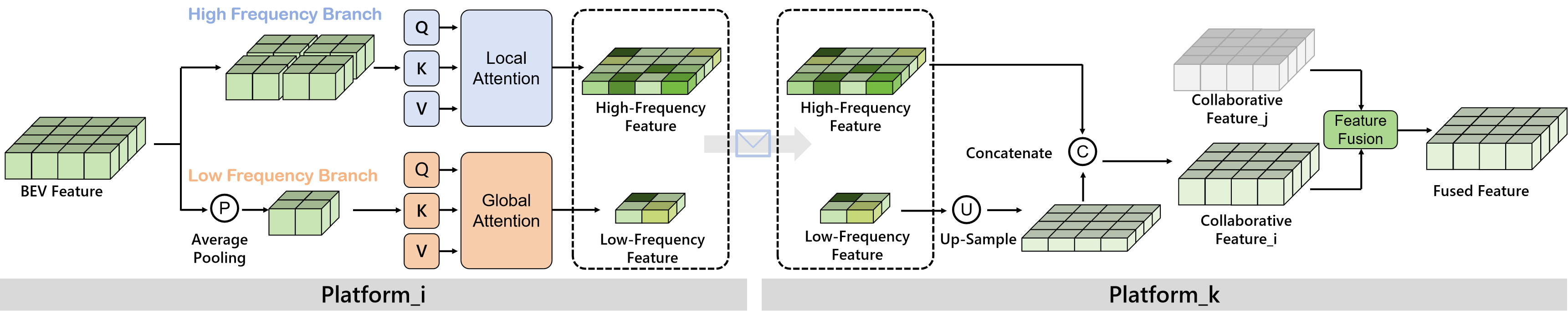} 
    \caption{Operation flow of the HLFDC module. This module compresses the original local features through two branches: low frequency and high frequency. In the high-frequency branch, local attention is conducted to obtain high-frequency features reflecting objects' edges, shapes, and other details. These high-frequency features are then condensed into low-dimensional channels. Conversely, the low-frequency branch preserves the overall structural information through low-filtered average pooling and global attention. The high-frequency branch condenses spatial information into down-sampled representations.}  
    \label{hlfdc}  
\end{figure*}

This section presents RS-DFM, a well-designed collaborative perception framework for remote sensing platforms, which enables multiple platforms to share visual information, promoting more holistic perception. As depicted in Fig.\ref{overview}, RS-DFM comprises four components (i.e., feature extraction, GBG module, HLFDC module, and supports for multiple downstream tasks) to implement the entire collaboration process. We briefly introduce these components in the overview, while delving into the innovative designs of the GBG and HLFDC modules in Section \ref{gbg} and Section \ref{hlfdc}, respectively.

\subsection{Overview}
Typically, the RS-DFM framework encompasses the following components.

\textbf{Feature Extraction}.
Initially, each platform captures and processes its observations using a shared encoder to extract semantic information. We utilize the Swin-Tiny\cite{liu2021swin} backbone as each drone's encoder due to its low inference latency and optimized memory usage within the family of transformer-based backbones.

\textbf{GBG Module}.
For collaboration among platforms, image features need to be transformed from each pixel coordinates to BEV representations in the world coordinates. 
Based on the encoded image features, the GBG module provides a generalized depth estimation for various platforms at different angles and heights by leveraging the geometric prior.
Then, image features and their corresponding estimated depths are used to generate frustum features, 
which are then converted into voxel features with the help of the camera's intrinsic and extrinsic parameters. 
Ultimately, sum-pooling and high-dimensional reduction are applied to the voxel features to derive BEV representations, convenient for later collaboration.

\textbf{HLFDC Module}.
The HLFDC Module aims to balance enhanced perception with transmission cost in inter-platform collaboration. 
Given the characteristics of remote sensing images, this module achieves information compression by decoupling high-frequency and low-frequency information. High-frequency information, which captures fine details, is preserved, whereas low-frequency information, representing the global structure, is compressed. Following collaborative interactions, local features and transmitted compressed collaborative features are fused through pixel-wise weighting, facilitating subsequent downstream tasks.

\textbf{Supports for Multiple Downstream Tasks.}
The collaboration among multiple platforms aims to enrich their respective perceptual information and is not limited to a specific task. Consequently, RS-DFM supports various downstream tasks, including 3D object detection, BEV instance segmentation, and multi-object tracking. Specifically, the fused features generated by the HLFDC module are fed into downstream task heads to obtain enhanced perception results.

\subsection{Generalized BEV Generation Module}
\label{sec:GBG}
Various remote sensing platforms lead to the significant discrepancy in observations of identical objects due to different viewing angles and altitudes.
Consequently, it’s necessary to convert multi-view observations into a unified spatial coordinate before information fusion.
To this end, we transform the 2D image features to the BEV coordinate system.
The BEV system offers distinct advantages, particularly in terms of scale and spatial consistency, which can effectively mitigate gaps among multi-view observations.
Depth estimation serves as a critical component in generating BEV representations. 
However, the vast range of aerial observations, coupled with various estimation intervals, presents challenges for the prevalent lss method in approximating the depth of each pixel.
To address this issue, we leverage the geometric priors provided by inclined observations and propose a novel Generalized BEV Generation Module. The procedural methodology of the proposed module is outlined below.

\subsubsection{Derivation of geometric constraint}
In this part, we utilize the camera's intrinsics and extrinsics parameters to calculate the corresponding depth upper-bound for each position in the observation.

According to the pinhole camera model, the transformation between the point $(x, y, z)$ in the world coordinate system and its projection $(u, v)$ in the pixel coordinate system can be described by the equation:
\begin{equation}
	\small
	\begin{bmatrix}
		x \\
		y \\
		z
	\end{bmatrix} = \mathbf{R}^{-1} \left( \mathbf{K}^{-1} \begin{bmatrix}
		u \\
		v \\
		1
	\end{bmatrix} {d} - \mathbf{T} \right).
\end{equation}
Here, the intrinsic matrix of the camera is denoted as $\mathbf{K} \in \mathbb{R}^{3 \times 3}$, and the extrinsics are represented by a rotation matrix $\mathbf{R} \in \mathbb{R}^{3 \times 3}$ and a translation vector $\mathbf{T} \in \mathbb{R}^{3 \times 1}$. The symbol $d$ represents the exact pixel depth in the camera coordinate system.

For notational brevity, we denote \( \mathbf{R}^{-1}\mathbf{K}^{-1} \) as the matrix \( \mathbf{M} = (m_{ij}) \in \mathbb{R}^{3 \times 3} \) and \( \mathbf{R}^{-1}(-\mathbf{T}) \) as the vector \( \mathbf{N} = (n_{i}) \in \mathbb{R}^{3 \times 1} \).
Hence, the equation (1) is modified as:
\begin{equation}
\small
	\begin{bmatrix}
		x \\
		y \\
		z
	\end{bmatrix} = \mathbf{M} \begin{bmatrix}
		u \\
		v \\
		1
	\end{bmatrix} d + \mathbf{N} = \begin{bmatrix}
		m_{11} & m_{12} & m_{13} \\
		m_{21} & m_{22} & m_{23} \\
		m_{31} & m_{32} & m_{33}
	\end{bmatrix} \begin{bmatrix}
		u \\
		v \\
		1
	\end{bmatrix} d + \begin{bmatrix}
		n_1 \\
		n_2 \\
		n_3
	\end{bmatrix}.
\end{equation}

In reference to the world coordinate system, the ground plane is located in the negative direction of the drone's y-axis. 
By setting \( y = -H \) in the equation (2), where \( H \) represents the altitude of the drone, 
the depth upper-bound of each position in the camera coordinate can be derived as:
\begin{equation}
\small
	{D_{(u,v)}}=\frac{-H-n_2}{m_{21} u+m_{22} v+m_{23}}.
\end{equation}
Following the acquisition of the depth upper bound for each pixel, we utilize this geometric constraint as a prior for optimizing the depth estimation.

\subsubsection{A relative depth estimation}
In contrast to the previous absolute depth estimation, we introduce a relative depth estimation method to calculate the ratio of the each pixel's depth to the depth upper-bound. 
The global unified depth estimation interval is replaced by a flexible relative depth estimation interval $(0, 1)$ scaled by the depth upper-bound for each region.

Specifically, we utilize a parametric network $\Phi_{\text{depth}}(\cdot)$, which generates a pixel-wise categorical distribution of depth, represented as $d_{\text{pred}} = \Phi_{\text{depth}}(F_{(u,v)}) \cdot D_{(u,v)} \in \mathbb{R}^{D}$. Here, $D$ signifies the number of discretized probability of depth bins.
We treat the estimation as a classification problem to capture the inherent depth estimation uncertainty and reduce the impact of erroneous depth estimates, as stated in LSS.

\subsubsection{BEV Generation}
The following process transforms 2D observations into BEV representations based on the aforementioned pinhole camera model, in equation (1). First, operate the outer product on the encoded features, $F_{\text{2D}} \in \mathbb{R}^{W \times H \times C}$, and the corresponding estimated depth, $D_{\text{pred}} = \left[ d_{\text{pred}_{(h,w)}} \right]_{(w \in W, h \in H)} \in \mathbb{R}^{W \times H \times D}$, to obtain the frustum features, which aim to recover the spatial information of the image. Then, project the frustum features into BEV pillars using the camera’s intrinsics and extrinsics. This operation converts multi-view observations into a unified world coordinate system. Finally, perform z-axis sum pooling at each spatial position $(x,y,:)$, collapsing the height dimension to generate the BEV features, $F_{\text{BEV}} \in \mathbb{R}^{X \times Y \times C}$. The obtained BEV representations facilitate subsequent information exchange among different platforms.

\subsection{High-Low Frequency Decoupled Collaboration Module}
\label{sec:HLFDC}
Communication among remote sensing platforms is often restricted, necessitating the partial transmission of valuable information to achieve efficient collaborative perception. As indicated in the literature\cite{Cooley_Lewis_Welch, Deng_Cahill_2005}, the high-frequency components of feature maps depict object details, while low-frequency components typically capture the global structure.
Given that remote sensing images contain small and dense foreground objects with rich background information, we tend to preserve intricate details and compress global spatial information for lightweight transmission. Therefore, we propose a collaborative strategy based on high-low frequency decoupling, implemented as follows:

\subsubsection{High Frequency Information Extraction}
The high-frequency branch is utilized to capture local dependencies that reflect fine details.
To this end, we implement a local window self-attention with the simple non-overlapping window partition.

The original BEV feature map is segmented into multiple local feature maps using non-overlapping windows of size $M \times M$. Each local feature map is then reshaped into a feature sequence, $F_{\text{seq}} \in \mathbb{R}^{M^2 \times C}$, which serves as the input to the multi-head self-attention layer.
Each self-attention head calculates the query $\mathbf{Q}$, key $\mathbf{K}$ and value $\mathbf{V}$ matrices with a linear transformation from $F_{\text{seq}}$,
\begin{equation}
	\mathbf{Q} = F_{\text{seq}}\mathbf{W}_q,  \mathbf{K} = F_{\text{seq}}\mathbf{W}_k,  \mathbf{V} = F_{\text{seq}}\mathbf{W}_v, \in \mathbb{R}^{M^2 \times D_h}
\end{equation}
where $\mathbf{W}_q$, $\mathbf{W}_k$, $\mathbf{W}_v \in \mathbb{R}^{C \times D_h}$ are learnable parameters and $D_h$ is the number of hidden dimensions for a head. Next, the output of a self-attention head is a weighted sum over $M^2$ value vectors,
\begin{equation}
	\mathrm{SA}_h(F_{\text{seq}}) = \mathrm{Softmax}\left( \frac{\mathbf{Q}\mathbf{K}^\top}{\sqrt{D_h}} \right) \mathbf{V}\in \mathbb{R}^{M^2 \times D_h}.
\end{equation}
For the multi-head self-attention layer with $N_h$ heads, the final output is computed by a linear projection of the concatenated outputs from each self-attention head, which can be formulated by:
\begin{equation}
	\mathrm{MSA}(F_{\text{Seq}}) = \mathrm{concat}_{h \in [N_h]} [\mathrm{SA}_h(F_{\text{Seq}})] \mathbf{W}_o \in \mathbb{R}^{M^2 \times C'},
\end{equation}
where $\mathbf{W}_o\in \mathbb{R}^{(N_h \times D_h) \times C'}$ is a learnable parameter. 
Herein, we set $C' = C/2$ and restructure the high-frequency feature sequences into their spatial format, which are then combined to form the complete high-frequency spatial feature $F_\text{high} \in \mathbb{R}^{X \times Y \times C'}$.
\subsubsection{Low Frequency Information Extraction}
Several studies\cite{Park_Kim} have indicated the effectiveness of the global attention mechanism in capturing low-frequency information.
However, due to the limited computational budget, it is expensive to directly perform global attention on original high-resolution feature maps.
Inspired by \cite{Voigtman_Winefordner_1986}, we first initially utilize an average pooling operation to achieve dual functions of low-pass filtering and down-sampling the feature maps. 
Considering the complementary high-low frequency, the stride of the average pooling is set to match the size of the local window in the high-frequency branch.

Specifically, the original BEV feature map is average-pooled with a kernel and stride of size $M$ to obtain the initial low-frequency signals, which are projected into $N_h$ queries, keys, and values, $\mathbf{Q}, \mathbf{K}, \mathbf{V} \in \mathbb{R}^{XY/M^2 \times D_h}$, respectively. Then, we apply the global multi-head self-attention to capture the rich low-frequency feature sequences that are finally reshaped to the low-frequency spatial features  $F_\text{low} \in \mathbb{R}^{X/M \times Y/M \times C'}$. The process of multi-head self-attention is same with the equations (4--6).
\subsubsection{Inter-platform Collaborative Features Fusion}
After obtaining the high-low frequency decoupled features at each remote sensing platform, these features are transmitted to other platforms to facilitate collaborative perception. 
Upon receiving the collaborative features, a geometric transformation $G$ is applied to align them with the local coordinate system, ensuring spatial congruence. 
Subsequently, the platform integrates high-frequency and low-frequency information to restore the original information. 
As the low-frequency features are average pooled, a series of transposed convolutions is utilized to up-sample them to the original size, ensuring the consistence with the shape of the high-frequency features.
Consequently, the processed high-low frequency features are concatenated to generate the collaborative features $F_\text{col}$ for subsequent fusion, which can be expressed as:
\begin{equation}
	F_\text{col} = \text{concat}[\text{Up}(G(F_\text{low})); G(F_\text{high})] \in \mathbb{R}^{X \times Y \times C}.
\end{equation}

Guided by local features \(F_{\text{BEV}}\), the contribution weight \(W_j\) of  $F_\text{col,j}$ towards constructing fused features $F_{\text{fused}}$ is quantified by:
\begin{equation}
	W_j = \frac{\phi_{\text{conv}}\left([F_{\text{BEV}}; F_\text{col,j}]\right)}{\sum_{j=1}^N \phi_{\text{conv}}\left([F_{\text{BEV}}; F_\text{col,j}]\right)} \in \mathbb{R}^{X \times Y \times 1}
\end{equation}
, where $\phi_\text{conv}$ represents the multi-layer convolution operations. Moreover, a pixel-level weighted fusion is executed to generate the fused features for subsequent downstream tasks: \begin{equation}
	F_{\text{fused}} = \sum_{j=1}^N W_j F_\text{col,j} \in \mathbb{R}^{X \times Y \times C}.
\end{equation}

\section{Experiments}
\subsection{Datasets}
\begin{figure*}
	\centering
	\includegraphics[width=1\textwidth]{"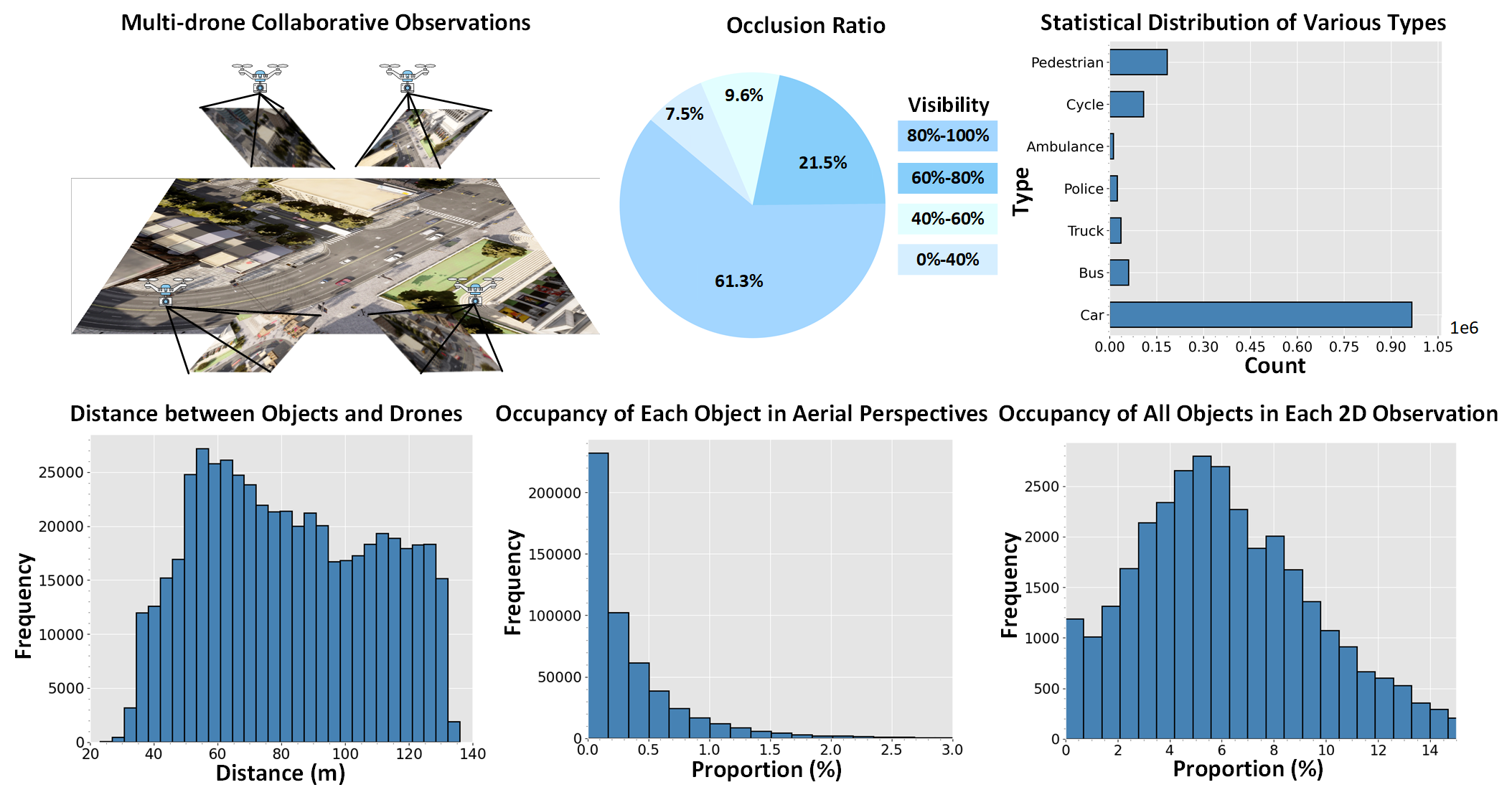"}
	\caption{Statistical charts for the AirCo-MultiTasks dataset, depicting the occlusion within a single view, the number of various object types, the distribution of distances between objects and drones, and the proportion of objects within observations, respectively.}
	\label{dataset}
\end{figure*}

Through collaboration, UAV swarms can achieve extended perception range, enhanced robustness, and improved flexibility, leading to better performance in various tasks under complex environments. However, available datasets for multi-drone collaboration primarily focus on detection and segmentation, lacking support for prediction tasks. To provide a more diverse and challenging benchmark for the multi-drone perception community, we introduce AirCo-MultiTasks, a simulation dataset based on CARLA\cite{dosovitskiy2017carla} that supports multi-task collaborative perception for multi-UAVs.

Specifically, five collaborative drones are positioned at intersections to monitor traffic flow from different directions. Three of these drones fly at a low altitude of 30 meters, one flies at 50 meters, and one flies at a high altitude of 70 meters, covering an area of approximately 150m $\times$ 150m. To illustrate the challenges of aerial observations intuitively, we present several charts to reflect the characteristics of AirCo-MultiTasks, such as occlusion, long-distant observations, small objects, etc., as shown in Fig.\ref{dataset}.

Each UAV is equipped with three types of sensors: a device for capturing RGB images, a device for environmental perception, and a device for transmitting perceptual information between UAVs. This setup enables the UAVs to achieve collaborative perception for various tasks, including 3D object detection, BEV semantic segmentation, and target trajectory prediction.

The dataset comprises 38,000 synchronous images with a resolution of 1600 x 900, captured by five collaborative drones. It is divided into 170 training scenes and 30 validation scenes. To facilitate multi-task training and evaluation, each frame is meticulously annotated with both 2D and 3D labels, encompassing three primary object categories: vehicles, bicycles, and pedestrians. This dataset serves as a valuable tool for assessing the performance of collaborative perception methods across a spectrum of perception tasks, including 2D and 3D object detection, BEV segmentation, and trajectory prediction. A detailed breakdown of the dataset is provided below.

\textbf{Data collection.} The proposed dataset was collected using CARLA simulation. We generated complex traffic scenarios in CARLA and deployed five UAVs at three different flight heights at intersections or roadsides to jointly monitor traffic flow in the same area from different perspectives.

\textbf{Sensor setup.} Each UAV is equipped with an RGB optical camera, and UAVs at different heights observe the same area with different pitch angles. The field of view (FoV) of each camera is 70°, and the resolution is 1600 $\times$ 900 pixels. The translation (x, y, z) and rotation (w, x, y, z in quaternion) of each camera are recorded during data collection. With this sensor setup and flight height configuration, the collaborative region observed by the five UAVs can cover approximately 150m $\times$ 150m.

\textbf{Scene creation.} Roads, static buildings, and dynamic targets in the simulation scene are all created by CARLA. We selected 33 locations for data collection from three high-quality simulation maps provided by CARLA. To enhance the richness and complexity of the scenes, the weather and traffic flow in each scene were randomly initialized during data collection.

\textbf{Traffic flow generation.} Vehicles and pedestrians in the simulation scene are controlled and managed by CARLA. The hundreds of dynamic targets in the scene inherit the Actor class in CARLA and are uniformly spawned, managed, and destroyed by the simulator. The number and location of target spawn points in each map are predetermined, and the spawn points are all located near roads.

\textbf{Data annotation.} We collected synchronous images from all cameras on five UAVs, brings about a sample size of 5 images. Camera intrinsics and extrinsics in global coordinates were provided to facilitate coordinate transformation across the UAVs. During data collection, 3D bounding boxes of vehicles and pedestrians were synchronously recorded, including location (x, y, z), rotation (w, x, y, z in quaternion) in the global coordinate system, and length, width, height, and visibility.

\subsection{Evaluation Metrics}

    
    
    
    

\subsubsection{Object Detection}
The 3D object detection evaluation metrics are mean Average Precision (mAP), mean Average Translation Error (mATE), mean Absolute Scale Error (mASE), and mean Absolute Orientation Error (mAOE). These metrics are defined as follows:

\begin{itemize}
\item \textbf{mAP}: This metric measures the average precision of the detected objects across different classes and various match thresholds. It uses the Average Precision (AP) metric, which defines a match by thresholding the 2D center distance on the ground plane instead of intersection over union. AP is calculated as the normalized area under the precision-recall curve, considering only points where recall and precision are both above 10\%. This approach helps to minimize the impact of noise often found in low precision and recall regions. Then, an average is computed over matching thresholds $D = \{0.5, 1, 2, 4\}$ meters and the set of classes $C$:
\begin{equation}
    \small
    \operatorname{mAP} = \frac{1}{|C||D|} \sum_{c \in C} \sum_{d \in D} AP_{c,d},
\end{equation}
where $AP_{c,d}$ is the Average Precision for class $c$ at threshold $d$.

    \item \textbf{mATE}: This metric evaluates the average translation error of the detection bounding box centers. It measures the accuracy of the model in detecting the position of the objects. The specific calculation is as follows:
    \begin{equation}
        \small
        \operatorname{mATE} = \frac{1}{N} \sum_{i=1}^{N} \|c_i - \hat{c}_i\|,
    \end{equation}
    where $c_i$ and $\hat{c}_i$ are the ground truth and predicted centers of the $i$-th bounding box, respectively, and $N$ is the total number of detected objects.
    
    \item \textbf{mASE}: This metric measures the average scale error of the detection bounding boxes. It evaluates the accuracy of the model in estimating the size of the objects. The specific calculation is as follows:
    \begin{equation}
        \small
        \operatorname{mASE} = \frac{1}{N} \sum_{i=1}^{N} \left|s_i - \hat{s}_i\right|,
    \end{equation}
    where $s_i$ and $\hat{s}_i$ are the ground truth and predicted scales (e.g., width, height, length) of the $i$-th bounding box, respectively, and $N$ is the total number of detected objects.
    
    \item \textbf{mAOE}: This metric assesses the average orientation error of the detection bounding boxes. It evaluates the accuracy of the model in estimating the orientation of the objects. The specific calculation is as follows:
    \begin{equation}
        \small
        \operatorname{mAOE} = \frac{1}{N} \sum_{i=1}^{N} \left|\theta_i - \hat{\theta}_i\right|,
    \end{equation}
    where $\theta_i$ and $\hat{\theta}_i$ are the ground truth and predicted orientations of the $i$-th bounding box, respectively, and $N$ is the total number of detected objects.
    
\end{itemize}

\subsubsection{Instance Segmentation}
We adopt four metrics widely used in previous works~\cite{de2022rethinking, fernando2023towards}, including Intersection-over-Union (IoU), Segmentation Quality (SQ), Recognition Quality (RQ), and Panoptic Quality (PQ). These metrics are defined as follows:

\begin{itemize}
    \item \textbf{IoU}: This metric evaluates the segmentation quality of objects. It is calculated as follows:
    \begin{equation}
        \small
        \operatorname{IoU}\left(\hat{y}^{\mathrm{seg}}, y^{\mathrm{seg}}\right)= \frac{\sum_{h, w} \hat{y}^{\mathrm{seg}} \cdot y^{\mathrm{seg}}}{\sum_{h, w} \hat{y}^{\mathrm{seg}}+y^{\mathrm{seg}}-\hat{y}^{\mathrm{seg}} \cdot y^{\mathrm{seg}}},
    \end{equation}
    where $\hat{y}^{\mathrm{seg}}$ and $y^{\mathrm{seg}}$ denote the predicted and ground truth semantic segmentation, respectively.

    \item \textbf{SQ}: This metric evaluates the segmentation quality of the correctly detected segments. It is the average IoU of the true positives (TP):
    \begin{equation}
        \small
        \operatorname{SQ}\left(\hat{y}^{\text{inst}}, y^{\text{inst}}\right)= \frac{\sum_{\left(p, q\right) \in TP} \operatorname{IoU}\left(p, q\right)}{|TP|}.
    \end{equation}

    \item \textbf{RQ}: This metric evaluates the recognition quality of the predicted segments, considering both true positives (TP) and the penalties for false positives (FP) and false negatives (FN):
    \begin{equation}
        \small
        \operatorname{RQ}\left(\hat{y}^{\text{inst}}, y^{\text{inst}}\right)= \frac{|TP|}{|TP|+\frac{1}{2}|FP|+\frac{1}{2}|FN|}.
    \end{equation}

    \item \textbf{PQ}: This metric considers both the quality of segmentation and the accuracy of instance recognition, capturing both the overlap between predicted and ground truth segments and the correct classification of instances. The metric is expressed as:
    \begin{equation}
        \small
        \operatorname{PQ}\left(\hat{y}^{\text{inst}}, y^{\text{inst}}\right)= \frac{\sum_{\left(p, q\right) \in TP} \operatorname{IoU}\left(p, q\right)}{\left|TP\right|+\frac{1}{2}\left|FP\right|+\frac{1}{2}\left|FN\right|}.
    \end{equation}
    $TP$, $FP$, and $FN$ correspond to true positives, false positives, and false negatives, respectively. $TP$ are the correctly predicted segments with an IoU over 0.5 and matching ground truth segments. $FP$ are the predicted segments that do not match any ground truth segments. $FN$ are the ground truth segments that are not predicted by the model.
\end{itemize}

\subsubsection{Trajectory Prediction}
We adopt IoU and Video Panoptic Quality (VPQ)~\cite{li2023powerbev, fernando2023towards} as the temporal evaluation metrics.

\begin{itemize}
    \item \textbf{IoU}: For frame-level evaluation, IoU evaluates the segmentation quality of objects at the present and future frames. The specific calculation is as follows:
    \begin{equation}
        \small
        \operatorname{IoU}\left(\hat{y}_t^{\mathrm{seg}}, y_t^{\mathrm{seg}}\right)=\frac{1}{N} \sum_{t=0}^{N-1} \frac{\sum_{h, w} \hat{y}_t^{\mathrm{seg}} \cdot y_t^{\mathrm{seg}}}{\sum_{h, w} \hat{y}_t^{\mathrm{seg}}+y_t^{\mathrm{seg}}-\hat{y}_t^{\mathrm{seg}} \cdot y_t^{\mathrm{seg}}},
    \end{equation}
    where $N$ denotes the number of output frames, $\hat{y}_t^{\mathrm{seg}}$ and $y_t^{\mathrm{seg}}$ denote the predicted and ground truth semantic segmentation at timestamp $t$, respectively.

    \item \textbf{VPQ}: For video-level evaluation, VPQ reflects the quality of the segmentation and ID consistency of the instances through the video, expressed as:
    \begin{equation}
        \small
        \operatorname{VPQ}\left(\hat{y}_t^{\text{inst}}, y_t^{\text{inst}}\right)=\sum_{t=0}^{N-1} \frac{\sum_{\left(p_t, q_t\right) \in TP_t} \operatorname{IoU}\left(p_t, q_t\right)}{\left|TP_t\right|+\frac{1}{2}\left|FP_t\right|+\frac{1}{2}\left|FN_t\right|},
    \end{equation}
    where $TP_t$, $FP_t$, and $FN_t$ correspond to true positives, false positives, and false negatives at timestamp $t$, respectively.
\end{itemize}

\subsection{Implementation Details}
We adhere to the setups introduced in previous studies~\cite{huang2021bevdet, hu2021fiery, li2023powerbev} for various BEV-specific downstream tasks. Initially, the raw images, with a resolution of $900 \times 1600$ pixels, are scaled and randomly cropped to $256 \times 704$ pixels. For view transformation, the relative depth estimation range is set from 0 to 1 and discretized into 100 intervals. For BEV representations, the spatial ranges for the $x$, $y$, and $z$ axes are configured as $[-75, 75]$ meters, $[-75, 75]$ meters, and $[0, 10]$ meters, respectively. Model performance is evaluated in two perceptual scopes: a $150 \, \text{m} \times 150 \, \text{m}$ area (long-range) and a $75 \, \text{m} \times 75 \, \text{m}$ area (short-range), both with a 0.75-meter resolution. Specifically, 
we adopt the backbone of Swin-Tiny~\cite{liu2021swin} as the encoder for each drone because of its light-weight design and low inference latency.
For the HLFDC module, the size of local window is set to 4.
Regard downstream tasks, BEVDET\cite{huang2021bevdet} is employed as the detector for 3D object detection, Fiery\cite{hu2021fiery} as the segmentator for BEV instance segmentation, and PowerBEV\cite{li2023powerbev} as the trajectory predictor for BEV trajectory prediction. Our proposed framework is trained using the AdamW optimizer with an initial learning rate of $2 \times 10^{-4}$. The training process is conducted on four NVIDIA A40 GPUs, with a batch size of 8 for 20 epochs.

\subsection{Comparative Baselines for Collaborative perception}

\textbf{Early collaboration} involves the integration of raw observations, where multi-view images are combined to generate BEV representations.

\textbf{Intermediate collaboration} focuses on feature-level interactions to produce comprehensive BEV representations. These approaches can be either fully connected or partially connected:

\begin{itemize}
    \item The fully connected paradigm shares complete features among all members. 
    \begin{itemize}
        \item V2VNet~\cite{wang2020v2vnet} performs multi-round message exchanges using graph neural networks.
        \item V2X-ViT~\cite{xu2022v2x} explores correlations among heterogeneous collaborators through transformer blocks.
    \end{itemize}
    \item The partially connected paradigm restricts interactions to certain members or regions. 
    \begin{itemize}
        \item Who2com~\cite{liu2020who2com} introduces a handshake mechanism to select collaborative partners.
        \item When2com~\cite{liu2020when2com} determines the optimal timing for initiating collaboration.
        \item Where2comm~\cite{hu2022where2comm} uses the detection head to direct interactions towards specific regions.
    \end{itemize}
\end{itemize}

\textbf{Late collaboration} combines individual prediction results from multiple drones to form a final decision.

\subsection{Quantitative Evaluation}
\subsubsection{Benchmark Comparison in 3D Object Detection}
Tab. \ref{table.detection} shows that our proposed approach, RS-DFM, achieves the highest performance in terms of mAP and mATE among all collaborative perception methods, even much better than the early collaboration without incorporating generalized feature mapping.
This superior performance can be attributed to our designed GBG module, which facilitates more accurate object localization and yields outstanding results in position-related metrics.
While our RS-DFM may exhibit slightly lower performance in mASE and mAOE, it still sharply reduces these errors compared to the majority of collaboration approaches. 
Additionally, RS-DFM significantly outperforms No-Collaboration, with a 202\% increase in mAP for the short range and a 254\% increase in mAP for the long range, indicating the effectiveness of collaboration. 
In contrast to fully connected baselines, our RS-DFM enhances object detection accuracy and recall by 2--4\% and reduces errors in localization, scale, and orientation by 6--30\%, while requiring only approximately half the transmission ratio. When compared to the partially connected baselines, RS-DFM surpasses the previous state-of-the-art method, Where2com, by a considerable margin, including a 4.4\% increase in mAP and decreases of 6\% in mATE, 10\% in mASE, and 9.5\% in mAOE during the long-range setting.

\begin{table*}[h!t]
	\renewcommand{\arraystretch}{1.1}
	\caption{A comparison of collaboration baselines for 3D object detection. 
	}
	\centering
  \resizebox{\textwidth}{!}{
			\begin{tabular}{cc|cc|cc|cc|cc}
				\toprule
				\multicolumn{2}{c|}{\multirow{2}{*}{\textbf{Methods}}} &   \multicolumn{2}{c|}{\textbf{mAP$\uparrow$}} & \multicolumn{2}{c|}{\textbf{mATE $\downarrow$}} & \multicolumn{2}{c|}{\textbf{mASE $\downarrow$}} & \multicolumn{2}{c}{\textbf{mAOE $\downarrow$}} \\
				\multicolumn{2}{c|}{} & \textbf{Short} & \textbf{Long} & \textbf{Short} & \textbf{Long} & \textbf{Short} & \textbf{Long} & \textbf{Short} & \textbf{Long}
				\\ \midrule
				\multicolumn{2}{c|}{No Collaboration} & 0.219 & 0.125 & 0.570 & 0.631 & 0.195 & 0.204 & 0.182 & 0.190 \\ \cline{1-10}
				\multicolumn{2}{c|}{Early Collaboration} & 0.641 & 0.437 & 0.382 & 0.430 & \textbf{0.160} & \textbf{0.169} & \textbf{0.065} & \textbf{0.083}\\ \cline{1-10}
				\multicolumn{2}{c|}{Late Collaboration} & 0.506 & 0.303 & 0.416 & 0.423 & 0.212  & 0.213& 0.101 & 0.103\\ \cline{1-10}
				\multicolumn{1}{c|}{\multirow{2}{*}{\shortstack{Intermediate Collaboration \\ (Fully Connected)}}} & V2X-ViT~\cite{xu2022v2x} & 0.638 & 0.434 & 0.410 & 0.457 & 0.240 & 0.208 & 0.094 & 0.116\\
				\multicolumn{1}{c|}{} & V2VNet~\cite{wang2020v2vnet} & 0.631 & 0.433 & 0.403 & 0.453 & 0.184 & 0.190 & \textit{0.088} & \textit{0.108} \\ \cline{1-10}
				\multicolumn{1}{c|}{\multirow{4}{*}{\shortstack{Intermediate Collaboration\\ (Partially Connected)}}} & Who2com~\cite{liu2020who2com} & 0.461 & 0.291 & 0.522 & 0.580 & 0.188 & 0.196 & 0.116 & 0.135 \\
				\multicolumn{1}{c|}{} & When2com~\cite{liu2020when2com} & 0.137 & 0.063 & 0.495 & 0.513 &  0.185 & 0.189 & 0.222 & 0.237 \\
				\multicolumn{1}{c|}{} & Where2comm~\cite{hu2022where2comm} & 0.634 & 0.424 & 0.398 & 0.447 & 0.189 & 0.196 & 0.118 & 0.136 \\
				\multicolumn{1}{c|}{} & RS-DFM (Ours) & \textcolor{black}{\textbf{0.662}} & \textcolor{black}{\textbf{0.443}}  & \textcolor{black}{\textbf{0.374}} & \textcolor{black}{\textbf{0.413}}& 
				\textcolor{black}{{0.170}} & \textcolor{black}{{0.176}}  & \textcolor{black}{0.109} & \textcolor{black}{0.123}  \\ \bottomrule
			\end{tabular}
	}
	\label{table.detection}
\end{table*}

\subsubsection{Benchmark Comparison in BEV Instance Segmentation}
As presented in Tab. \ref{table.segmentation}, our method demonstrates the top performance in instance segmentation across all evaluation metrics among collaborative approaches. Compared with the second-best early collaboration approach, which processes multi-view raw observations in a centralized manner, our RS-DFM improves the IoU by 2.8--3.1\% and PQ by 2.4--5.1\% through feature-level interactions,  facilitated by our designed GBG and HLFDC modules.
Unlike the intricate task of 3D object detection, which considers orientation, instance segmentation focuses on recognizing and segmenting objects to reflect recall and localization accuracy. 
Attributed to our GBG module improving feature mapping from 2D to BEV, our RS-DFM surpasses prevalent fully connected collaboration approaches by 3--4.5\% in IoU and 1.8--5\% in PQ.
In contrast to previous sparse collaborative perception methods that reduce transmission costs at the expense of performance, our method outperforms these baselines by 35--75\% in IoU and 42--63\% in PQ.
\begin{table*}[tb!]
	\renewcommand{\arraystretch}{1.1}
	\caption{A comparison of collaboration baselines for BEV instance segmentation.
	}
	\centering
 \resizebox{\textwidth}{!}{
			\begin{tabular}{cc|cc|cc|cc|cc}
				\toprule
				\multicolumn{2}{c|}{\multirow{2}{*}{\textbf{Models}}} &   \multicolumn{2}{c|}{\textbf{IoU (\%) $\uparrow$}} & \multicolumn{2}{c|}{\textbf{PQ (\%) $\uparrow$}} & \multicolumn{2}{c|}{\textbf{SQ (\%) $\uparrow$}} & \multicolumn{2}{c}{\textbf{RQ (\%) $\uparrow$}} \\
				\multicolumn{2}{c|}{} & \textbf{Short} & \textbf{Long} & \textbf{Short} & \textbf{Long} & \textbf{Short} & \textbf{Long} & \textbf{Short} & \textbf{Long}
				\\ \midrule
				\multicolumn{2}{c|}{No Collaboration} & 37.02 & 18.74 & 31.59 & 17.06 & 72.14 & 71.60 & 43.78 & 23.82 \\ \cline{1-10}
				\multicolumn{2}{c|}{Early Collaboration} & 62.15 & 29.66 & 49.23 & 26.67 & 74.13 & 73.63 & 66.41 & 36.22\\ \cline{1-10}
				\multicolumn{2}{c|}{Late Collaboration} & 53.66 & 26.10 & 31.12 & 17.41 & 67.65 & 67.54 & 46.01 & 25.78\\ \cline{1-10}
				\multicolumn{1}{c|}{\multirow{2}{*}{\shortstack{Intermediate Collaboration \\ (Fully Connected)}}} & V2X-ViT~\cite{xu2022v2x} & 62.10 & 29.84 & 49.74 & 26.49 & 73.87 & 73.19 & 67.33 & 36.20\\
				\multicolumn{1}{c|}{} & V2VNet~\cite{wang2020v2vnet} & 61.68 & 30.25 & 48.60 & 26.83 & 73.47 & 72.98 & 66.15 & 36.77 \\ \cline{1-10}
				\multicolumn{1}{c|}{\multirow{4}{*}{\shortstack{Intermediate Collaboration\\ (Partially Connected)}}} & Who2com~\cite{liu2020who2com} & 47.10 & 21.73 & 35.89 & 18.76 & 70.54 & 70.41 & 50.88 & 26.65 \\
				\multicolumn{1}{c|}{} & When2com~\cite{liu2020when2com} & 36.45 & 18.40 & 31.17 & 16.85 & 71.34 & 70.73 & 43.69 & 23.83 \\
				\multicolumn{1}{c|}{} & Where2comm~\cite{hu2022where2comm} & 51.63 & 25.69 & 35.60 & 19.68 & 67.75 & 67.50 & 52.54 & 29.15 \\
				\multicolumn{1}{c|}{} & RS-DFM (Ours) & \textcolor{black}{\textbf{63.92}} & \textcolor{black}{\textbf{31.17}}  & \textcolor{black}{\textbf{51.04}} & \textcolor{black}{\textbf{27.30}}& 
				\textcolor{black}{\textbf{75.47}} & \textcolor{black}{\textbf{74.76}}  & \textcolor{black}{\textbf{67.63}} & \textcolor{black}{\textbf{36.51}}   \\ \bottomrule
			\end{tabular}
	}
	\label{table.segmentation}
\end{table*}

\subsubsection{Benchmark Comparison in Trajectory Prediction}
As presented in Tab. \ref{table.prediction}, it is challenging to predict objects’ future trajectories solely based on single-view observation, and the performance of future instance segmentation and motion prediction significantly improves with the help of any collaborative methods. 
Distinct from observation patterns of single-frame instance segmentation, previous collaborative approaches, such as early collaboration, V2XViT, and V2VNet, exhibit comparative performance with our method in the short range. This suggests that temporal information can mitigate the estimation error of feature mapping.
However, the gap between our method and previous collaborative methods becomes evident in the long-range setting. 
Our RS-DFM outperforms early collaboration by 2.5\% in IoU and 2\% in VPQ. Compared to fully connected collaborative approaches, our method surpasses V2X-ViT by 3.8\% in IoU and 5.7\% in VPQ, and V2VNet by 17.8\% in IoU and 12.8\% in VPQ.
Notably, even with the same transmission ratio as the partially connected baseline Where2comm, our method significantly outperforms it by 10.8\% in IoU and by 42.4\% in VPQ, showcasing the effectiveness of our HLDFC module.
\begin{table}[tb!]
	\renewcommand{\arraystretch}{1.5}
	\caption{A comparison of collaboration baselines for trajectory prediction.
	}
	\setlength{\tabcolsep}{2.5pt}
	\centering
	\resizebox{0.5\textwidth}{!}{
		\begin{tabular}{cc|cc|cc}
			\toprule
			\multicolumn{2}{c|}{\multirow{2}{*}{\textbf{Models}}} & \multicolumn{2}{c|}{\textbf{IoU (\%) $\uparrow$}} & \multicolumn{2}{c}{\textbf{VPQ (\%) $\uparrow$}} \\
			\multicolumn{2}{c|}{} & \textbf{Short} & \textbf{Long} & \textbf{Short} & \textbf{Long} \\ \midrule
			\multicolumn{2}{c|}{No Collaboration} & 25.05 & 13.43 & 11.51 & 5.92 \\ \cline{1-6}
			\multicolumn{2}{c|}{Early Collaboration} & 60.70 & 34.77 & \textbf{50.07} & 29.70 \\ \cline{1-6}
			\multicolumn{2}{c|}{Late Collaboration} & 58.23 & 31.48 & 46.42 & 25.91 \\ \cline{1-6}
			\multicolumn{1}{c|}{\multirow{2}{*}{\shortstack{Intermediate Collaboration \\ (Fully Connected)}}} & V2X-ViT~\cite{xu2022v2x} & 59.74 & 34.32 & 49.44 & 28.64 \\
			\multicolumn{1}{c|}{} & V2VNet~\cite{wang2020v2vnet} & 60.68 & 30.25 & 48.60 & 26.83 \\ \cline{1-6}
			\multicolumn{1}{c|}{\multirow{4}{*}{\shortstack{Intermediate Collaboration \\ (Partially Connected)}}} & Who2com~\cite{liu2020who2com} & 50.26 & 27.65 & 40.75 & 27.65 \\
			\multicolumn{1}{c|}{} & When2com~\cite{liu2020when2com} & 45.41 & 25.66 & 36.26 & 16.85 \\
			\multicolumn{1}{c|}{} & Where2comm~\cite{hu2022where2comm} & 58.18 & 32.18 & 26.20 & 21.26 \\
			\multicolumn{1}{c|}{} & RS-DFM (Ours) & \textbf{61.02} & \textbf{35.65} & {50.05} & \textbf{30.27} \\ \bottomrule
		\end{tabular}
	}
	\label{table.prediction}
\end{table}

\subsection{Ablation Studies}
\subsubsection{Effectiveness of Proposed Modules}
Our RS-DFM collaborative framework introduces two innovative components: the GBG and HLFDC modules. We evaluate these modules based on their ability to enhance downstream tasks in the long-range setting, including 3D object detection, BEV instance segmentation, and trajectory prediction, as well as their ability to optimize the performance-transmission trade-off, as depicted in Tab. \ref{modules.ablation}.
The GBG-only variant significantly enhances 3D object detection accuracy, with improvements of approximately 4\% in mAP and a reduction in position deviation by about 7.2\% in mATE. This is primarily attributable to the more accurate BEV representations guided by the generalized feature mapping. Notably, while the GBG module improves mAP, it may lead to orientation and scale errors, as observed in the benchmark comparisons mentioned previously. This phenomenon merits further investigation.
For instance segmentation, the GBG-only variant improves segmentation and recognition quality by 1.5-5.4\%. In trajectory prediction, this variant also shows improvements of 3.5\% in IoU and 2.6\% in VPQ. The evaluation metrics for these tasks are highly related to position precision, thereby demonstrating the effectiveness of our designed GBG module.
On the other hand, the HLFDC-only variant reduces transmission costs by 43.8\%, with only a marginal performance decrease of less than 1\% in mAP, IoU, and VPQ relative to the baseline. 
This indicates the validity of the high-low frequency decoupled collaboration, albeit with minor performance trade-offs.
Overall, our RS-DFM, integrated with both GBG and HLFDC modules, achieves a balance between collaborative perception and transmission efficiency.
\begin{table*}[ht]
	\centering
	\caption{The ablation study of proposed modules. 'Ratio' in the table refers to the transmission ratio during collaborative perception.}
	\label{table.3}
	\renewcommand{\arraystretch}{1.3}
	\scalebox{1.1}{\setlength{\tabcolsep}{0.9mm}{
			\begin{tabular}{@{}cc|cccc|cccc|cc|c@{}}
				\toprule
				\multirow{2}{*}{\textbf{GBG}} & \multirow{2}{*}{\textbf{HLFDC}} & \multicolumn{4}{c|}{\textbf{3D Object Detection}} & \multicolumn{4}{c|}{\textbf{BEV Instance Segmentation}} & \multicolumn{2}{c|}{\textbf{Trajectory Prediction}} & \multirow{2}{*}{\textbf{Ratio}} \\  
				& & \textbf{mAP $\uparrow$} & \textbf{mATE $\downarrow$} & \textbf{mASE $\downarrow$} & \textbf{mAOE $\downarrow$} & \textbf{IoU (\%) $\uparrow$} & \textbf{PQ (\%) $\uparrow$} & \textbf{SQ (\%) $\uparrow$ } & \textbf{RQ (\%) $\uparrow$} & \textbf{IoU (\%) $\uparrow$} & \textbf{VPQ (\%) $\uparrow$} \\ \midrule
				$\times$ & $\times$ & 0.437 & 0.430 & 0.169 & 0.083 & 29.66 & 26.67 & 73.63 & 36.22 & 34.77 & 29.70 & 1 \\
				\checkmark & $\times$ & 0.455 & 0.399 & 0.172 & 0.135 & 31.24 & 27.65 & 75.24 & 36.74 & 35.97 & 30.46 & 1 \\
				$\times$ & \checkmark & 0.433 & 0.443 & 0.189 & 0.105 & 29.64 & 26.05 & 72.81 & 35.77 & 34.59 & 29.35 & 0.563 \\
				\checkmark & \checkmark & 0.443 & 0.413 & 0.176 & 0.123 & 31.17 & 27.30 & 74.76 & 36.51 & 35.65 & 30.27 & 0.563 \\ \bottomrule
			\end{tabular}
	}}
\label{modules.ablation}
\end{table*}
\subsubsection{Effect of window size in HLFDC module}
\begin{table}[tb!]
	\renewcommand{\arraystretch}{1.4}
	\caption{A comparison of performance metrics for different window sizes implemented in the HLFDC module.}
	\setlength{\tabcolsep}{2.5pt}
	\centering
	\resizebox{0.45\textwidth}{!}{
		\begin{tabular}{c|cccccc}
			\toprule
			\textbf{Window Size} & \textbf{1} & \textbf{2} & \textbf{4} & \textbf{8} & \textbf{16} & \textbf{32} \\ \midrule
			{Det-mAP} & 0.458 & 0.447 & 0.443 & 0.432 & 0.425 & 0.402 \\ \cline{1-7}
			{Seg-IoU} & 31.21 & 30.79 & 31.17 & 30.58 & 30.23 & 30.05 \\ \cline{1-7}
			{Pred-IoU} & 35.78 & 35.60 & 35.65 & 35.34 & 34.90 & 34.73 \\ \cline{1-7}
			{Ratio} & 1.000 & 0.625 & 0.531 & 0.508 & 0.502 & 0.500 \\ \bottomrule
		\end{tabular}
	}
	\label{table.window_size}
\end{table}
The window size determines the contextual range in each local window and the compression ratio for low-frequency information. 
Smaller windows are more sensitive to capturing high-frequency local feature variations but may be susceptible to noise, and result in a considerable transmission expense for collaborative interactions. Conversely, larger windows encompass a broader range of textual information within the local windows but may potentially lose significant information for low-frequency global attention processing. After investigating the effects of various window sizes across different downstream tasks, as shown in Table \ref{table.window_size}, we select a 4$\times$4 window size in the HLFDC module for optimal performance.

\subsection{Visualization }
\subsubsection{Collaborative 3D object detection}
Fig.\ref{det} illustrates that collaborative perception significantly enhances the detection of occluded and out-of-range objects through inter-platform interactions compared with no collaboration. 
For instance, UAV1 hardly recognizes any objects due to significant occlusion in its single observations.
However, collaborative approaches successfully recall these occluded objects by leveraging complementary information from other platforms. 
Additionally, this collaboration alleviates the challenge of detecting small objects at high altitudes, as demonstrated by the row of UAV5.
Intuitively, our method exhibits superior detection performance compared to other approaches. In particular, in the red circles in the rows corresponding to UAV3, UAV4, and UAV5, early collaboration and Where2comm recall numerous erroneous bounding boxes, whereas our method significantly reduces such errors. This improvement is attributed to our specially designed GBG module, which achieves better feature mapping by incorporating geometric priors.
\subsubsection{Collaborative BEV instance segmentation}
\begin{figure*}[h]
	\centering
	\begin{minipage}[b]{\textwidth}
		\centering
		\includegraphics[width=\textwidth]{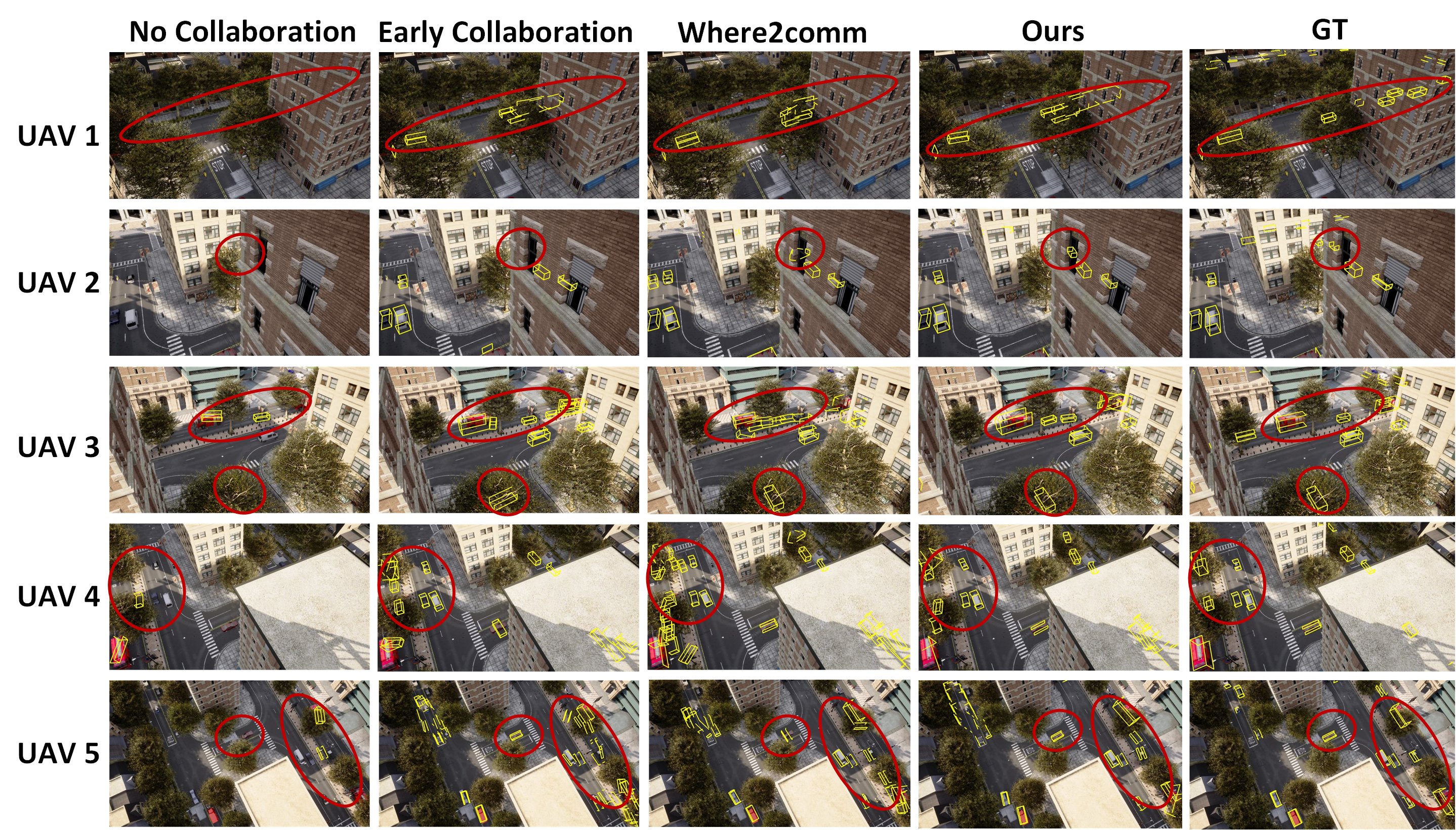}
		\captionof{figure}{Visualization of collaborative 3D object detection. Red circles highlight the areas where other baselines make mistaken predictions.}
		\label{det}
	\end{minipage}
	
	\vspace{0.5cm}
	
	\begin{minipage}[b]{\textwidth}
		\centering
		\includegraphics[width=\textwidth]{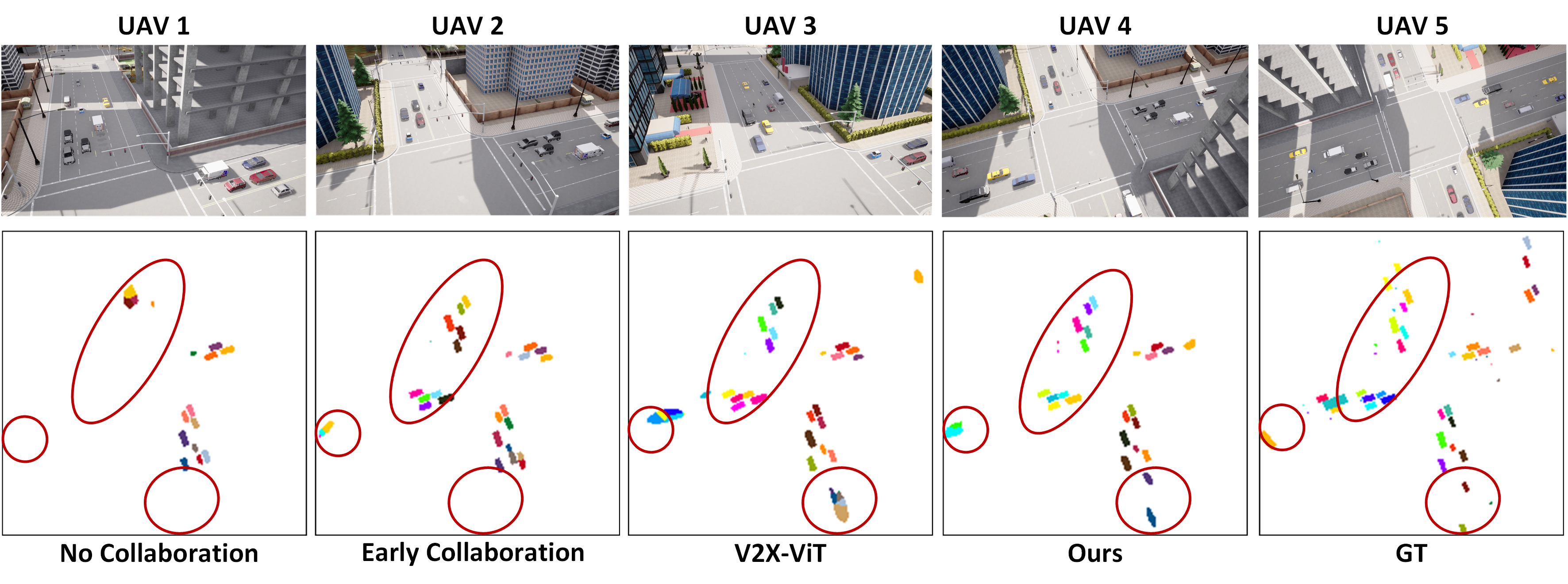}
		\captionof{figure}{Visualization of collaborative instance segmentation. Each instance is allocated a distinct color. Red circles highlight the areas where other baselines make mistaken predictions.}
		\label{seg}
	\end{minipage}
	
	\vspace{0.5cm}
	
	\begin{minipage}[b]{\textwidth}
		\centering
		\includegraphics[width=\textwidth]{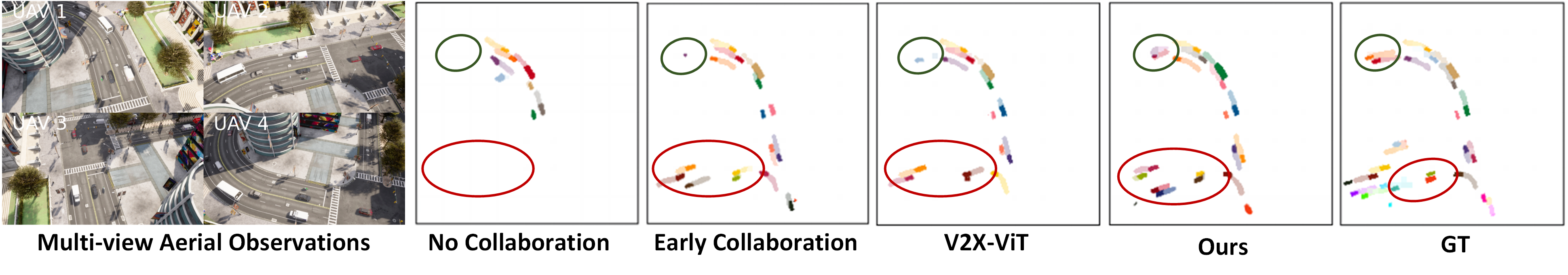}
		\captionof{figure}{Visualization of collaborative trajectory prediction. Each instance is allocated a distinct color, and its predicted trajectory is represented with the same color and slight transparency. Red circles highlight the areas where other baselines make mistaken predictions.}
		\label{pred}
	\end{minipage}
\end{figure*}

The task of BEV instance segmentation focuses on segmenting the BEV representation into small object bricks, where accuracy highly relies on precise mapping from 2D observations to the BEV. Our method is obviously better at accurately segmenting distant objects, typically positioned at the edges of the BEV, as depicted by the two small red circles in Figure \ref{seg}. While early collaboration can centralize the processing of raw observations, it still performs less effectively than our approach, which leverages sparse feature-level interactions. This suggests that precise feature mapping holds more significance than transmitting a larger quantity of features.
\subsubsection{Collaborative trajectory prediction}
Trajectory prediction heavily depends on inter-frame continuity. Previous sparse collaborative approaches often convey spatially discrete features, potentially disrupting this continuity. However, our method maintains the overall structure of intermediate BEV features and extracts critical information through high and low-frequency decoupling. As a result, our method achieves comparable performance to early collaboration and fully-connected intermediate collaboration paradigms.
Leveraging our specially designed GBG and HLFDC modules, our method accurately forecasts the trajectories of multiple objects in challenging intersections and achieves more precise segmentation and prediction results than other well-known baselines. These findings are consistent with our quantitative evaluations.

\section{Conclusion}
We proposed a novel distributed remote sensing foundation model, RS-DFM. The core idea is to utilize the geometric priors of oblique perspectives to generalize and map features from different spatial observations into a unified feature space, facilitating subsequent multi-task perception. Simultaneously, we decouple high-frequency and low-frequency features in remote sensing and perform selective information compression to minimize the loss of critical information during compression. Comprehensive experiments covering multi-altitude, multi-perspective inputs, and multi-task outputs demonstrate that RS-DFM achieves an excellent balance between perception performance and communication bandwidth in multi-platform collaborative online inference.

\bibliographystyle{ieeetr}
\bibliography{reference}

\end{document}